\documentclass[runningheads]{llncs}
\usepackage[T1]{fontenc}
\usepackage{graphicx}
\usepackage{booktabs}
\usepackage{fancyhdr}
\usepackage{amsmath}
\usepackage{indentfirst}
\usepackage{listings}
\usepackage{sectsty}
\usepackage{color, colortbl}
\usepackage{amssymb}
\usepackage{thmtools}
\usepackage{shadethm}
\usepackage{hyperref}
\usepackage{tablefootnote}
\usepackage{threeparttable}
\usepackage{setspace}
\usepackage{tikz}
\usepackage{adjustbox}
\usepackage{graphicx}
\usepackage{xcolor}
\usepackage{multirow}
\usepackage{twemojis}
\usepackage{marvosym}

\newcommand\extrafootertext[1]{%
    \bgroup
    \renewcommand\thefootnote{\fnsymbol{footnote}}%
    \renewcommand\thempfootnote{\fnsymbol{mpfootnote}}%
    \footnotetext[0]{#1}%
    \egroup
}
\definecolor{LightCyan}{rgb}{0.88,1,1}
\DeclareMathOperator*{\argmax}{arg\,max}
\begin{document}
%
\title{HistGen: Histopathology Report Generation via Local-Global Feature Encoding and Cross-modal Context Interaction}
%
%
\author{Zhengrui Guo\inst{1} \and
Jiabo Ma\inst{1} \and
Yingxue Xu\inst{1} \and
Yihui Wang\inst{1} \and
Liansheng Wang\inst{3} \and
Hao Chen\inst{1,2 (\twemoji{envelope})}}
%
\authorrunning{Zhengrui Guo et al.}
%
\institute{$^1$Department of Computer Science and Engineering, The Hong Kong University of Science and Technology \\ \email{\{zguobc, jmabq, yxueb, ywangrm\}@connect.ust.hk, jhc@cse.ust.hk} \\ $^2$Department of Chemical and Biological Engineering, The Hong Kong University of Science and Technology \\ $^3$School of Information Science and Engineering, Xiamen University \\ \email{lswang@xmu.edu.cn}}
%
\maketitle              
\begin{abstract}
Histopathology serves as the gold standard in cancer diagnosis, with clinical reports being vital in interpreting and understanding this process, guiding cancer treatment and patient care. The automation of histopathology report generation with deep learning stands to significantly enhance clinical efficiency and lessen the labor-intensive, time-consuming burden on pathologists in report writing. In pursuit of this advancement, we introduce \textbf{HistGen}, a multiple instance learning-empowered framework for histopathology report generation together with the first benchmark dataset for evaluation. Inspired by diagnostic and report-writing workflows, HistGen features two delicately designed modules, aiming to boost report generation by aligning whole slide images (WSIs) and diagnostic reports from local and global granularity. To achieve this, a local-global hierarchical encoder is developed for efficient visual feature aggregation from a region-to-slide perspective. Meanwhile, a cross-modal context module is proposed to explicitly facilitate alignment and interaction between distinct modalities, effectively bridging the gap between the extensive visual sequences of WSIs and corresponding highly summarized reports. Experimental results on WSI report generation show the proposed model outperforms state-of-the-art (SOTA) models by a large margin. Moreover, the results of fine-tuning our model on cancer subtyping and survival analysis tasks further demonstrate superior performance compared to SOTA methods, showcasing strong transfer learning capability. Dataset, model weights, and source code are available in \url{https://github.com/dddavid4real/HistGen}. 
\end{abstract}
\extrafootertext{$^{\text{\twemoji{envelope}}}$ Corresponding Author.}
\section{Introduction}
\noindent Histopathology tissue analysis reveals critical details about tumor characteristics and is considered the gold standard in cancer diagnosis and prognosis \cite{khened2021generalized}. 
In the last decade, the increasing availability of digital WSIs has given rise to computational pathology (CPath), promoting diagnostics with computer-assisted tools.
Recent success of deep learning has notably advanced this field, outperforming experienced pathologists on specific tasks \cite{campanella2019clinical,zhang2022dtfd,shao2021transmil}.
However, pathologists are still required to compile findings and craft reports for each WSI, a task that is labor-intensive, time-consuming, and error-prone.
Thus, the automated generation of pathology reports holds immense potential in facilitating the report-writing process and alleviating the heavy workloads for pathologists. Additionally, these automated reports can underscore abnormalities and offer diagnostic rationales, aiding pathologists in their analyses.

Contrasting with the field of radiology report generation \cite{chen2020generating,chen2022cross,nicolson2023improving}, WSI report generation faces unique challenges. A primary obstacle is the lack of recognized benchmark datasets; existing datasets predominantly focus on patch-level images and captions \cite{gamper2021multiple,huang2023visual,lu2023visual}, neglecting comprehensive, global descriptions of entire WSIs. 
Additionally, the gigapixel size of WSIs poses unique challenges for the visual encoding stage of report generation and hinders the direct adoption of existing report generation methods. 
Moreover, the contrast between enormous informative WSI patches of dense visual signals and succinctly summarized diagnostic reports of discrete textual tokens underscores the need for cross-modal interactions.
Further, prevalent WSI analysis methods rely on multiple instance learning (MIL), necessitating pre-trained feature extractors, which we empirically identify as a critical bottleneck, impeding optimal performance in WSI report generation.

\noindent \textbf{Related Work:} 

\noindent\textit{Multiple instance learning in CPath.} Given the gigapixel size of WSIs and GPU memory limitations \cite{araujo2019computing}, MIL framework is prevalently used for WSI analysis. This method tiles a WSI into patches, using a pre-trained feature extractor to convert them into feature vectors, which are then aggregated into a WSI-level representation for prediction \cite{campanella2019clinical,feng2017deep,ilse2018attention,shao2021transmil,li2021dual,zhang2022dtfd}. 
Several works \cite{feng2017deep,campanella2019clinical} used max-pooling to locate the most suspicious instances for slide-level prediction. To engage information of all patches, a side network was employed in \cite{ilse2018attention} to compute attention score of each patch and use it as the weight for aggregation. Li et al. \cite{li2021dual} utilized cosine distance between instances as the weighting factor for aggregation. Shao et al. \cite{shao2021transmil} further adopted transformer layers to explore the long-range instance correlations. 
However, recent studies in MIL mainly focus on WSI-level prediction tasks, with WSI report generation under-explored along this research direction, which is the focus of this work.

\noindent \textit{Image captioning and report generation.} Image captioning \cite{vinyals2015show,anderson2018bottom,vaswani2017attention,cornia2020meshed} and radiology report generation \cite{chen2020generating,chen2022cross} are the most relevant fields to this work. These tasks require models to generate descriptive captions or diagnostic reports for input natural or radiology images. 
Various studies \cite{vinyals2015show,anderson2018bottom} adopted the CNN-RNN framework for image encoding and caption decoding. Recent advent of Transformer \cite{vaswani2017attention} enabled capturing long-range dependencies, with further explorations proposed to bridge encoding and decoding stages \cite{cornia2020meshed}. This concept is similarly applied in radiology report generation \cite{chen2020generating,chen2022cross}.
However, the gigapixel size of WSIs requires immense computational capacity, making the direct application of these methods to WSI data impractical. Pioneering efforts in WSI report generation have been made \cite{sengupta2023automatic,guevara2023caption,chen2023mi}, yet they mainly involve simple combinations of existing MIL and captioning models (e.g., \cite{chen2023mi} combined TransMIL \cite{shao2021transmil} with a transformer decoder), neglecting the connections between visual and textual modalities along with their distinct characteristics.


\noindent \textbf{Contribution:} 

\noindent Building on the success of image captioning \cite{vinyals2015show,anderson2018bottom,cornia2020meshed,wang2023controllable} and radiology report generation \cite{chen2020generating,chen2022cross,nicolson2023improving}, and addressing unique challenges in WSI data, we present \textbf{HistGen}, a MIL-based histopathology report generation framework, designed to address WSI report generation from a cross-modal interaction perspective by aligning WSIs and diagnostic reports across local and global granularity.
Our primary contributions are summarized as follows:
\begin{itemize}
    \item[$\bullet$] We curate a benchmark WSI-report dataset of around 7,800 pairs by building a report collection and cleaning pipeline based on the TCGA platform \cite{tomczak2015review}.
    \item[$\bullet$] A local-global hierarchical visual encoder is proposed for effective encoding and aggregation of extensive WSI patch features in a region-to-slide manner. 
    \item[$\bullet$] We develop a cross-modal context module to enable interactions between visual encoding and textual decoding, leveraging cross-modal information. 
    \item[$\bullet$] Further, we pre-train a general-purpose MIL feature extractor on over 55,000 WSIs to boost MIL feature encoding.
    \item[$\bullet$] With extensive experimental validation, our framework outperforms SOTA methods on WSI report generation by a large margin. Moreover, experiment results on cancer subtyping and survival analysis tasks exhibit superior transfer learning capability of HistGen, with it achieving SOTA on both tasks.
\end{itemize} 

\section{Method}

\subsection{WSI-Report Dataset Curation}\label{sec:dataset}
\noindent Differing from the well-established benchmarks in radiology report generation \cite{johnson2019mimic,demner2016preparing}, there is a lack of accessible WSI-report datasets, hindering the advancement of this field.
In light of this, our goal is to curate a publicly available WSI-report dataset to serve as an evaluation benchmark. Based on uncurated report data from the TCGA platform \cite{tomczak2015review}, we build a dataset curation pipeline detailed as follows. We begin by downloading diagnostic report PDFs from TCGA and extracting raw texts from them. Since raw texts are still noisy and cluttered with redundant information, we utilize GPT-4 for further cleaning and summarization. By matching the case IDs from the reports with corresponding WSIs, we form a WSI-report dataset with 7,753 pairs, encompassing various diseases from different primary sites (see Tab. \ref{tab:dataset} in \textbf{Supplementary Materials} for more details). 

\subsection{HistGen for Automated WSI Report Generation}\label{sec:architecture}

\begin{figure}[htbp]
  \centering
  \includegraphics[width=\textwidth]{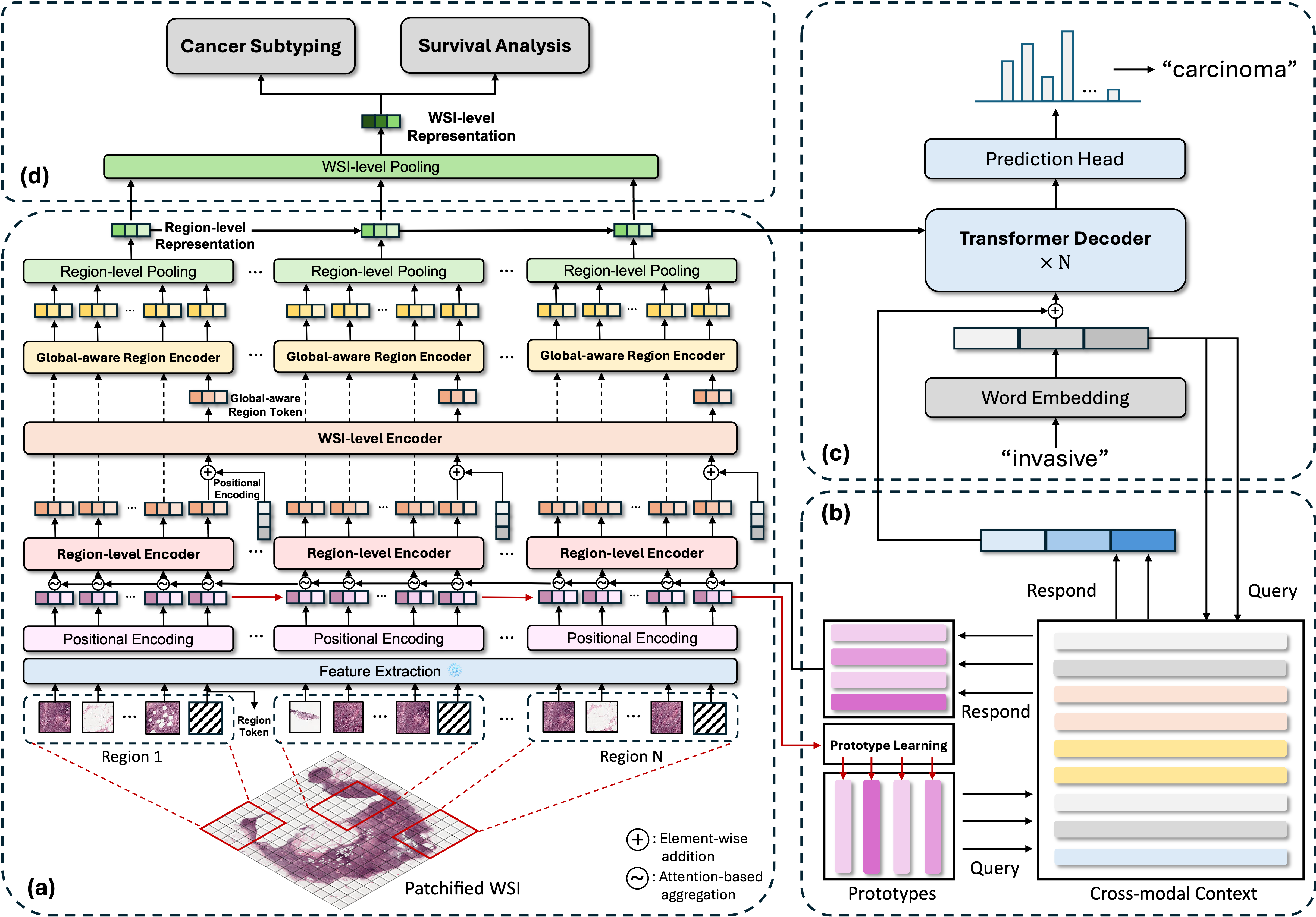}
  \caption{Overview of the proposed HistGen framework: (a) local-global hierarchical encoder module, (b) cross-modal context module, (c) decoder module, (d) transfer learning strategy for cancer diagnosis and prognosis.}
  \label{fig:overview}
\end{figure}

\noindent \textbf{Overview.} Fig. \ref{fig:overview} shows the overview of our HistGen framework. Fig. \ref{fig:overview}a denotes the local-global hierarchical visual encoder, which firstly extracts patch features with our pre-trained backbone and then efficiently encodes them under a local-to-global manner. After that, the visual features are fed into the decoder module (shown in Fig. \ref{fig:overview}c) for report generation. Fig. \ref{fig:overview}b corresponds to the cross-modal context-aware learning module, designed to bridge different modalities and store the knowledge from previous iterations for the model to refer to. 
Further, Fig. \ref{fig:overview}d highlights our strategy for fine-tuning the model to WSI-level prediction tasks.

\noindent\textbf{Pre-training feature extractor.} 
To boost the MIL feature encoding process, we collect over 55,000 WSIs from public datasets (see Fig. \ref{fig:WSI_distribution} and \ref{fig:patch_distribution} in \textbf{Supplementary Materials} for more details) to develop a general-purpose feature extractor that offers informative and robust patch embeddings. After tissue segmentation and patch tiling, we leverage around 200 million patches to pre-train a ViT-L model $f(\cdot;\theta)$ with DINOv2 strategy \cite{oquab2023dinov2}, which is then used for feature extraction, as depicted in Fig. \ref{fig:overview}a.

\noindent\textbf{Local-global hierarchical encoder (LGH).} 
Efficiently leveraging the information and correlations among WSIs' extensive patch sequence is a vital problem. Given that pathology reports typically contain both region-level details and overall WSI diagnoses, we deduce that the visual encoder should process visual features from a local-to-global perspective to effectively capture this granularity.

\noindent In this work, we propose the local-global hierarchical encoder (Fig. \ref{fig:overview}a) for region-aware representation learning. Specifically, we first segment the WSI into $N$ regions $X_i=\{x_{i,1}, x_{i,2}, x_{i,S}\}, i=1,\cdots,N$ with the region size $S$. Besides, we additionally add a region token $x_{i, S+1}$ at the end of each region sequence to learn a region-level representation. After that, we employ $f(\cdot;\theta)$ to obtain patch embeddings and add positional encoding (PE) to each region sequence:
\begin{equation*}
\begin{aligned}
    X_i' = \{f(x_{i,1};\theta) + e_{i,1}, \cdots, f(x_{i,S+1};\theta) + e_{i,S+1}\} = \{x_{i,1}', \cdots, x_{i,S+1}'\}.
\end{aligned}
\end{equation*}
Consequently, we design a region-level encoder $E_l(\cdot;\theta_l)$ to process $X_i'$. To capture global information and long-range dependencies among patches from various regions, we introduce PE to each region-level representation token $x_{i, S+1}'$, followed by processing through a WSI-level encoder $E_g(\cdot;\theta_g)$ for global interactions. Subsequently, these global-aware region tokens are reorganized to infuse global information into original regions with $E_l(\cdot;\theta)$. Finally, we obtain region-level representations $X_i''$ by attentive pooling on each region:
\begin{equation*}
\begin{aligned}
    X_i''&=\text{Pooling}\left(\{x_{i,1}'', \cdots, x_{i,S+1}''\}\right)=\text{Pooling}\left(E_l(x_{i,1}', \cdots, \Tilde{x}_{i,S+1}';\theta_l)\right) \\
    &=\text{Pooling}\left(E_l(x_{i,1}', \cdots, E_g(x_{1,S+1}',x_{2,S+1}',\cdots,x_{N,S+1}';\theta_g)_{i,S+1};\theta_l)\right),
\end{aligned}
\end{equation*}
and then input region representations $\{X_1'', X_2'', \cdots, X_N''\}$ into the decoder (see Fig. \ref{fig:overview}c) for report generation.
Moreover, to prove that our model learns diagnosis-related information during report generation, we further design a transfer learning mechanism (see Fig. \ref{fig:overview}d) by pooling the region-level representations $X_i''$ into a WSI-level representation $X''$ for downstream diagnosis and prognosis tasks.


\noindent\textbf{Cross-modal context module (CMC).} Given that report generation is an image-to-text task involving the transformation between distinct modalities, bridging the gap between pixel-based visual inputs and token-based textual outputs is crucial. Meanwhile, leveraging the model’s accumulated knowledge and memory from previous iterations is expected to significantly enhance this task.

\noindent The CMC module (Fig. \ref{fig:overview}b) is proposed in light of the above insights, denoted as $C = \{C_1, C_2, \cdots, C_m\}$, where $C_i$ is the context vector stored in this module. For visual inputs $\{X_1',\cdots, X_N'\}$, we select key patches $\{p_1, \cdots, p_l\}$ via cross-attention-based prototype learning, which aims to use representative patches as queries for the CMC module, avoiding redundancy from entire patch sequence. After querying, the responses generated by the CMC module are aggregated back into the original visual features, infusing them with cross-modal information.
This step aims to help the model leverage learned knowledge from the training data that might not be captured by the input embeddings alone. Similar interactions are applied to textual embeddings $\{y_1, y_2, \cdots, y_t\}$ from the decoder, except for prototype learning.
Additionally, the CMC module serves as an external memory, enabling the model to access and update iteratively, thereby progressively enhancing generation performance.

\noindent \textbf{Loss function.} Let $I=\{x_1, x_2, \cdots, x_n\}$ be a WSI with $n$ denoting the patch number and $T=\{y_1, y_2, \cdots, y_t\}$ be the corresponding reports with $t$ representing the number of tokens. The objective function is to maximize the conditional probability of the $T$ given $I$: $\theta^*=\argmax_{\theta}\sum_{i=1}^t\log P(y_i|y_1,y_2,\cdots,y_{i-1},I;\theta)$,
where $P(y_i|y_1,y_2,\cdots,y_{i-1},I)$ is the probability of generating the next word $y_i$ given the WSI $I$ and the preceding words $y_1,y_2,\cdots,y_{i-1}$ in the report.

\section{Experiments}

\subsection{Implementation Details}
\noindent \textbf{Datasets.} In this study, we focus on WSI report generation as well as cancer subtyping and survival analysis (split set as 80\%/10\%/10\% for train/val/test). For report generation, our model is evaluated on the WSI-report dataset (specified in Sec. \ref{sec:dataset}),
using NLG metrics including BLEU \cite{papineni2002bleu}, METEOR \cite{denkowski2011meteor}, and ROUGE-L \cite{lin2004rouge}. For cancer subtyping, since the TCGA reports may include phenotype-related information, we evaluate on three external classification datasets, including UBC-OCEAN\footnote{\url{https://www.kaggle.com/competitions/UBC-OCEAN}}, Camelyon \cite{bandi2018detection,bejnordi2017diagnostic}, and TUPAC16 \cite{veta2019predicting}, using Monte Carlo cross-validated Accuracy and AUC. For survival analysis, we evaluate on six datasets including BRCA, STAD, KIRC, KIRP, LUAD, and COADREAD from TCGA platform, with Monte Carlo cross-validated concordance index (c-Index). 


\noindent\textbf{Model settings.} For pre-trained DINOv2 ViT-L, the hidden dimension is 1,024 for feature extraction. For report generation model, we employ the LGH module with region size $96$, and a 3-layer Transformer decoder with 8 attention heads, and 512 dimensions for hidden states. For the cross-modal context module, we set it with dimension $512\times 2048$. Our model is trained under cross entropy loss with Adam optimizer, using learning rate $1\times 10^{-4}$ and weight decay by $0.8$ per epoch. We adopt beam search with a beam size of $3$ for inference.

\subsection{WSI Report Generation Results}
\noindent\textbf{Pre-trained feature extractor.} 
We compare our pre-trained DINOv2 ViT-L with ImageNet-pretrained ResNet50 and WSI-pretrained CTransPath \cite{wang2022transformer}. Most MIL methods utilize ImageNet-pretrained ResNet50, and results in Tab. \ref{tab:nlg} show that this leads to relatively low generation performance. We then turn to CTransPath \cite{wang2022transformer}, which is pre-trained on 30,000 WSIs using contrastive learning strategy. However, Tab. \ref{tab:nlg} demonstrates only minor increases in the generation performance of baseline models. In contrast, as illustrated in Tab. \ref{tab:nlg}, the baseline models show significant improvement by adopting our DINOv2 ViT-L, and several models such as R2GenCMN could already achieve plausible generation performance. This indicates that our DINOv2 ViT-L significantly outperforms both out-of-domain pretrained and domain-aligned pretrained methods, boosting subsequent tasks such as report generation.

\begingroup
\setlength{\tabcolsep}{6pt} 
\renewcommand{\arraystretch}{1.5} 
\begin{table}[h]
\centering
\caption{Comparisons of the proposed \textbf{HistGen} with other SOTA models on WSI report generation w.r.t NLG metrics.}
\begin{adjustbox}{width=\textwidth}
\scalebox{0.8}{
\begin{tabular}{c|lcccccc}
\toprule
\toprule
\multirow{2}{*}{\textbf{Feature Extractor}} & \multicolumn{1}{c}{\multirow{2}{*}{\textbf{Methods}}} & \multicolumn{6}{c}{\textbf{Metric}} \\ 
\cline{3-8}
& & BLEU-1 & BLEU-2 & BLEU-3 & BLEU-4 & METEOR & ROUGE-L \\ 
\midrule
\midrule
\multirow{6}{*}{\textbf{ResNet50}} 
& \textbf{Show\&Tell \cite{vinyals2015show}} & 0.249 & 0.099& 0.047& 0.025& 0.086& 0.165\\
\cline{2-8}
& \textbf{UpDownAttn \cite{anderson2018bottom}} & 0.250& 0.115& 0.065& 0.043& 0.096& 0.180\\
\cline{2-8}
& \textbf{Transformer \cite{vaswani2017attention}} & 0.249& 0.114& 0.065& 0.042& 0.095& 0.176\\
\cline{2-8}
& \textbf{M2Transformer \cite{cornia2020meshed}} & 0.250& 0.115& 0.065& 0.042& 0.095& 0.180\\
\cline{2-8}
& \textbf{R2Gen \cite{chen2020generating}} & 0.240& 0.105& 0.058& 0.036& 0.089& 0.177\\
\cline{2-8}
& \textbf{R2GenCMN \cite{chen2022cross}} & 0.225& 0.095& 0.047& 0.022& 0.094& 0.151\\
\midrule
\multirow{6}{*}{\textbf{CTransPath \cite{wang2022transformer}}} 
& \textbf{Show\&Tell \cite{vinyals2015show}} &0.262 &0.126 &0.071 &0.043 &0.094 &0.184 \\
\cline{2-8}
& \textbf{UpDownAttn \cite{anderson2018bottom}} & 0.240& 0.139& 0.090& 0.063& 0.100&0.201 \\
\cline{2-8}
& \textbf{Transformer \cite{vaswani2017attention}} & 0.271& 0.165&0.112 & 0.082&0.113 &0.227 \\
\cline{2-8}
& \textbf{M2Transformer \cite{cornia2020meshed}}  &0.259 &0.160 &0.108 &0.076 &0.103 &0.218 \\
\cline{2-8}
& \textbf{R2Gen \cite{chen2020generating}} & 0.237& 0.135& 0.085& 0.054& 0.086& 0.205\\
\cline{2-8}
& \textbf{R2GenCMN \cite{chen2022cross}} &0.211 &0.098 &0.054 &0.033 &0.079 &0.158 \\
\midrule
\multirow{6}{*}{\begin{tabular}[c]{@{}l@{}}\textbf{DINOv2 ViT-L} \\ ~~~~~~~\textbf{(Ours)}\end{tabular}}
& \textbf{Show\&Tell \cite{vinyals2015show}} & 0.189 & 0.094 & 0.056 & 0.039 & 0.070 & 0.165 \\
\cline{2-8}
& \textbf{UpDownAttn \cite{anderson2018bottom}} & 0.320 & 0.206 & 0.147 & 0.112 & 0.131 & 0.271 \\
\cline{2-8}
& \textbf{Transformer \cite{vaswani2017attention}} &0.382 &0.266 &0.200 &0.157 & 0.162 & 0.316 \\
\cline{2-8}
& \textbf{M2Transformer \cite{cornia2020meshed}} & 0.321 & 0.213& 0.152& 0.112& 0.131& 0.266 \\
\cline{2-8}
& \textbf{R2Gen \cite{chen2020generating}} &0.274 &0.166 &0.107 &0.071 &0.102 &0.234 \\
\cline{2-8}
& \textbf{R2GenCMN \cite{chen2022cross}} &0.381 &0.264 &0.199 &0.157 &0.164 &0.313 \\
\cline{2-8}
& \cellcolor{red!10!white} \textbf{Ours} & \cellcolor{red!10!white}\textbf{0.413}& \cellcolor{red!10!white}\textbf{0.297}& \cellcolor{red!10!white}\textbf{0.229}& \cellcolor{red!10!white}\textbf{0.184}& \cellcolor{red!10!white}\textbf{0.182}&\cellcolor{red!10!white}\textbf{0.344} \\
\bottomrule
\bottomrule
\end{tabular}}
\end{adjustbox}
\label{tab:nlg}
\end{table}
\endgroup

\noindent\textbf{WSI report generation results analysis.} For SOTA report generation methods in Tab. \ref{tab:nlg}, Show\&Tell \cite{vinyals2015show} and UpDownAttn \cite{anderson2018bottom} utilize CNN for image encoding and RNN for language decoding, manifesting lower performance compared to Transformer-based methods such as R2Gen \cite{chen2020generating}. M2Transformer \cite{cornia2020meshed} and R2GenCMN \cite{chen2022cross} further improve the performance by aligning encoding and decoding stages. 
However, visual token selection methods are needed to apply methods to WSI report generation due to the large sequence length, degrading report generation performance.
Our proposed method addresses this problem and outperforms these methods by a large margin via efficiently exploring the local and global information inside the patch sequence, interactively aligning the visual and textual context, and effectively utilizing the model's knowledge and memory from previous iterations. Building upon the DINOv2 ViT-L feature extractor, we compare our method to the aforementioned models. Tab. \ref{tab:nlg} confirms the superiority of our report generation model over SOTA models, with an average of around 3\% improvement observed on all NLG metrics (\textbf{Qualitative Analysis} is shown in \textbf{Supplementary Materials} Fig. \ref{fig:qualitative}).

\noindent\textbf{Ablation Studies.} We conduct additional experiments to determine the effects of our proposed modules.
Tab. \ref{tab:ablation1} illustrates the performance of baseline models, in which \textit{Base} represents a vanilla transformer with an additional pooling layer to accommodate the long sequence of WSI patches.
Our findings reveal that progressive stacking of the proposed modules leads to incremental improvements in NLG metrics, indicating that both the LGH and CMC modules play essential roles in boosting WSI report generation.

\begingroup
\setlength{\tabcolsep}{6pt} 
\renewcommand{\arraystretch}{1.5} 
\begin{table}[h]
\centering
\caption{Ablation study on the proposed LGH and CMC modules}
\begin{adjustbox}{width=\textwidth}
\scalebox{0.8}{
\begin{tabular}{lccccccc}
\toprule
\multirow{2}{*}{\textbf{Methods}}  & \multicolumn{7}{c}{\textbf{Metric}} \\ 
\cline{2-8}
& BLEU-1 & BLEU-2 & BLEU-3 & BLEU-4 & METEOR & ROUGE-L & AVG. $\Delta$\\ 
\midrule
\textbf{Base} & 0.384 & 0.267 & 0.203 & 0.163 & 0.166 & 0.316 & - \\
\hline
\textbf{~+ CMC} & 0.398 & 0.282 & 0.216 & 0.173 & 0.173 & 0.332 & \textbf{5.18\%} \\
\hline
\rowcolor{red!10!white}\textbf{~+ CMC + LGH} & \textbf{0.413}& \textbf{0.297}& \textbf{0.229}& \textbf{0.184}& \textbf{0.182 }&\textbf{0.344} & \textbf{10.50\%} \\
\bottomrule
\end{tabular}}
\end{adjustbox}
\label{tab:ablation1}
\end{table}
\endgroup

\subsection{Transfer Learning for Cancer Diagnosis and Prognosis}
\noindent WSI report generation could be considered as vision-language pre-training. To prove that our model learns diagnosis-related information during this task, we further fine-tune our model and evaluate on cancer subtyping and survival analysis with the strategy illustrated in Fig. \ref{fig:overview}d. 
Max-MIL, Mean-MIL, AB-MIL \cite{ilse2018attention}, TransMIL \cite{shao2021transmil}, DS-MIL \cite{li2021dual}, and DTFD-MIL \cite{zhang2022dtfd} are included for comparison.\\

\noindent\textbf{Cancer Subtyping.} Tab. \ref{tab:classification} summarizes the comparison of our model to SOTA MIL methods in cancer subtyping datasets, showing that our method outperforms all MIL approaches on all three classification tasks by a significant margin.

\begingroup
\setlength{\tabcolsep}{3pt} 
\renewcommand{\arraystretch}{1.4} 
\begin{table}[h]
\centering
\caption{Transfer Learning Results on \textbf{Cancer Subtyping} (measured with accuracy and AUC). \colorbox{red!10!white}{$\textbf{Mean}_{\textbf{Std}}$} represents the best performance and \colorbox{LightCyan}{$\textbf{Mean}_{\textbf{Std}}$} denotes the second best performance.}
\begin{adjustbox}{width=\textwidth}
\scalebox{0.8}{\begin{tabular}{l|cc|cc|cc|cc}
\toprule
\toprule
\multirow{2}{*}{\textbf{Methods}}  & \multicolumn{2}{c|}{\textbf{UBC-OCEAN}} &  \multicolumn{2}{c|}{\textbf{CAMELYON}} & \multicolumn{2}{c|}{\textbf{TUPAC16}} & \multicolumn{2}{c}{\textbf{Average}}\\ 
\cline{2-9}
& Acc & AUC  & Acc & AUC  & Acc & AUC & Acc & AUC  \\ 
\midrule
\midrule
\textbf{MaxMIL} & $0.767_{\pm 0.005}$&$0.940_{\pm 0.001}$ & $0.946_{\pm 0.001}$ & $0.979_{\pm 0.000}$ &$0.514_{\pm 0.002}$ &$0.693_{\pm 0.001}$ &$0.785_{\pm 0.002}$ &$0.894_{\pm 0.001}$\\
\hline
\textbf{MeanMIL} &$0.771_{\pm 0.004}$&$0.947_{\pm 0.001}$& $0.811_{\pm 0.002}$ &$0.898_{\pm 0.001}$ &$0.571_{\pm 0.002}$&$0.724_{\pm 0.005}$  &$0.738_{\pm 0.003}$ & $0.866_{\pm 0.002}$\\
\hline
\textbf{ABMIL \cite{ilse2018attention}} & \cellcolor{LightCyan}$0.792_{\pm 0.004}$&\cellcolor{LightCyan}$0.954_{\pm 0.000}$ & \cellcolor{LightCyan}$0.951_{\pm 0.000}$& \cellcolor{LightCyan}$0.988_{\pm 0.000}$ &\cellcolor{LightCyan}$0.573_{\pm 0.002}$&$0.736_{\pm 0.002}$  &\cellcolor{LightCyan}$0.809_{\pm 0.002}$&\cellcolor{LightCyan}$0.913_{\pm 0.001}$ \\
\hline
\textbf{TransMIL \cite{shao2021transmil}} &$0.782_{\pm 0.002}$ &$0.929_{\pm 0.001}$& $0.944_{\pm 0.001}$ & $0.976_{\pm 0.000}$ &$0.541_{\pm 0.004}$ &$0.689_{\pm 0.003}$ & $0.795_{\pm 0.002}$& $0.889_{\pm 0.001}$\\
\hline
\textbf{DS-MIL \cite{li2021dual}} &$0.785_{\pm 0.003}$&$0.944_{\pm 0.000}$ & $0.937_{\pm 0.002}$ &$0.979_{\pm 0.001}$ &$0.559_{\pm0.006}$ &$0.719_{\pm 0.002}$ &$0.797_{\pm 0.003}$ &$0.902_{\pm 0.001}$ \\
\hline
\textbf{DTFD-MIL \cite{zhang2022dtfd}} &$0.785_{\pm 0.004}$&\cellcolor{red!10!white}$0.955_{\pm 0.000}$ & $0.943_{\pm 0.014}$ & $0.978_{\pm 0.015}$&$0.555_{\pm 0.003}$ &\cellcolor{red!10!white}$0.739_{\pm 0.004}$ &$0.799_{\pm 0.008}$ &$0.910_{\pm 0.008}$\\
\hline
\hline
\textbf{Ours}  &\cellcolor{red!10!white}$0.808_{\pm 0.003}$  & $0.946_{\pm 0.001}$ & \cellcolor{red!10!white}$0.961_{\pm 0.000}$ & \cellcolor{red!10!white}$0.996_{\pm 0.000}$ &\cellcolor{red!10!white}$0.604_{\pm 0.002}$& \cellcolor{LightCyan}$0.738_{\pm 0.004}$ &\cellcolor{red!10!white}$0.827_{\pm 0.001}$&\cellcolor{red!10!white}$0.915_{\pm 0.001}$\\
\bottomrule
\bottomrule
\end{tabular}}
\end{adjustbox}
\label{tab:classification}
\end{table}
\endgroup

\begingroup
\setlength{\tabcolsep}{3pt} 
\renewcommand{\arraystretch}{1.4} 
\begin{table}[h]
\centering
\caption{Transfer Learning Results on \textbf{Survival Analysis} (measured with c-Index). \colorbox{red!10!white}{$\textbf{Mean}_{\textbf{Std}}$} represents the best performance and \colorbox{LightCyan}{$\textbf{Mean}_{\textbf{Std}}$} denotes the second best performance.}
\begin{adjustbox}{width=\textwidth}
\scalebox{0.8}{\begin{tabular}{l|ccccccc}
\toprule
\toprule
\textbf{Methods} & \textbf{ BRCA} &  \textbf{ STAD} & \textbf{ KIRC} & \textbf{ KIRP} & \textbf{ LUAD} & \textbf{ COADREAD} & \textbf{Average}\\ 
\midrule
\midrule
\textbf{MaxMIL} & $0.586_{\pm 0.010}$& $0.553_{\pm 0.006}$& \cellcolor{red!10!white}$0.732_{\pm 0.002}$ & $0.562_{\pm 0.022}$ & $0.563_{\pm 0.004}$ & $0.560_{\pm 0.012}$ & $0.595_{\pm 0.009}$ \\
\hline
\textbf{MeanMIL} & $0.608_{\pm 0.006}$ & $0.569_{\pm 0.005}$& \cellcolor{LightCyan}$0.710_{\pm 0.007}$& $0.609_{\pm 0.011}$ &$0.584_{\pm 0.005}$ & \cellcolor{red!10!white}$0.589_{\pm 0.011}$ & $0.612_{\pm 0.009}$ \\
\hline
\textbf{ABMIL \cite{ilse2018attention}} &$0.621_{\pm 0.005}$ &$0.574_{\pm 0.008}$ &$0.700_{\pm 0.009}$ & $0.636_{\pm 0.017}$ & \cellcolor{LightCyan}$0.603_{\pm 0.005}$& \cellcolor{LightCyan}$0.572_{\pm 0.006}$ & \cellcolor{LightCyan}$0.618_{\pm 0.008}$ \\
\hline
\textbf{TransMIL \cite{shao2021transmil}} & $0.615_{\pm 0.010}$ & $0.582_{\pm 0.003}$ & $0.672_{\pm 0.010}$ & $0.532_{\pm 0.017}$ & $0.584_{\pm 0.006}$ & $0.525_{\pm 0.008}$ & $0.593_{\pm 0.009}$\\
\hline
\textbf{DS-MIL \cite{li2021dual}} &\cellcolor{LightCyan}$0.624_{\pm 0.011}$  &$0.600_{\pm 0.006}$ & $0.672_{\pm 0.003}$& \cellcolor{LightCyan}$0.641_{\pm 0.013}$ &$0.591_{\pm 0.004}$ & $0.561_{\pm 0.006}$ & $0.593_{\pm 0.008}$ \\
\hline
\textbf{DTFD-MIL \cite{zhang2022dtfd}} &$0.592_{\pm 0.006}$ &\cellcolor{LightCyan}$0.601_{\pm 0.005}$ & $0.702_{\pm 0.008}$ &$0.572_{\pm 0.001}$ &$0.575_{\pm 0.005}$ &$0.561_{\pm 0.011}$ & $0.601_{\pm 0.006}$\\
\hline
\hline
\textbf{Ours}  & \cellcolor{red!10!white}$0.630_{\pm 0.008}$ & \cellcolor{red!10!white}$0.607_{\pm 0.003}$& $0.698_{\pm 0.006}$ & \cellcolor{red!10!white}$0.648_{\pm 0.015}$& \cellcolor{red!10!white}$0.623_{\pm 0.004}$ & $0.568_{\pm 0.011}$ & \cellcolor{red!10!white}$0.628_{\pm 0.008}$ \\
\bottomrule
\bottomrule
\end{tabular}}
\end{adjustbox}
\label{tab:survival}
\end{table}
\endgroup

\noindent\textbf{Survival Analysis.} Results of six survival analysis tasks, summarized in Tab. \ref{tab:survival}, reveal that our model outperforms others in most tasks and maintains the highest average score across all six datasets, manifesting superior transfer capabilities.\\

\noindent Tab. \ref{tab:classification} and \ref{tab:survival} highlight the effectiveness of our proposed model on WSI-level prediction tasks as well as the feasibility of transferring our model to these downstream tasks, confirming its ability to capture diagnostic information during WSI report generation.

\section{Conclusion}
\noindent In this work, we introduce \textbf{HistGen}, a MIL-empowered framework designed to enhance automated histopathology report generation, covering from model design to benchmark evaluation. To this end, 
we develop an advanced WSI report generation model by efficiently leveraging local-region and global-WSI information together with explicitly aligning the cross-modal encoding and decoding stages. To conduct experimental evaluation, we curate a benchmark WSI-report dataset, containing around 7,800 WSI-report pairs with high text quality. Extensive experiments on report generation, cancer subtyping, and survival analysis demonstrate the superiority of the proposed model as well as its strong capabilities of transfer learning. While this work lays the groundwork for WSI report generation, its scope is currently limited to histopathology. Future efforts will expand to include other fields such as radiology and ophthalmology, treating report generation across different fields as a unified problem within a single framework.


\clearpage
%
%
%
\bibliographystyle{splncs04}
\bibliography{samplepaper}

\begin{thebibliography}{10}
\providecommand{\url}[1]{\texttt{#1}}
\providecommand{\urlprefix}{URL }
\providecommand{\doi}[1]{https://doi.org/#1}

\bibitem{anderson2018bottom}
Anderson, P., He, X., Buehler, C., Teney, D., Johnson, M., Gould, S., Zhang, L.: Bottom-up and top-down attention for image captioning and visual question answering. In: Proc. IEEE Conf. Comput. Vis. Pattern Recogn. pp. 6077--6086 (2018)

\bibitem{araujo2019computing}
Araujo, A., Norris, W., Sim, J.: Computing receptive fields of convolutional neural networks. Distill  \textbf{4}(11), ~e21 (2019)

\bibitem{bandi2018detection}
Bandi, P., Geessink, O., Manson, Q., Van~Dijk, M., Balkenhol, M., Hermsen, M., Bejnordi, B.E., Lee, B., Paeng, K., Zhong, A., et~al.: From detection of individual metastases to classification of lymph node status at the patient level: the camelyon17 challenge. IEEE Trans. Med. Imaging  \textbf{38}(2),  550--560 (2018)

\bibitem{bejnordi2017diagnostic}
Bejnordi, B.E., Veta, M., Van~Diest, P.J., Van~Ginneken, B., Karssemeijer, N., Litjens, G., Van Der~Laak, J.A., Hermsen, M., Manson, Q.F., Balkenhol, M., et~al.: Diagnostic assessment of deep learning algorithms for detection of lymph node metastases in women with breast cancer. Jama  \textbf{318}(22),  2199--2210 (2017)

\bibitem{campanella2019clinical}
Campanella, G., Hanna, M.G., Geneslaw, L., Miraflor, A., Werneck Krauss~Silva, V., Busam, K.J., Brogi, E., Reuter, V.E., Klimstra, D.S., Fuchs, T.J.: Clinical-grade computational pathology using weakly supervised deep learning on whole slide images. Nat. Med.  \textbf{25}(8),  1301--1309 (2019)

\bibitem{chen2023mi}
Chen, P., Li, H., Zhu, C., Zheng, S., Yang, L.: Mi-gen: Multiple instance generation of pathology reports for gigapixel whole-slide images. arXiv preprint arXiv:2311.16480  (2023)

\bibitem{chen2022cross}
Chen, Z., Shen, Y., Song, Y., Wan, X.: Cross-modal memory networks for radiology report generation. arXiv preprint arXiv:2204.13258  (2022)

\bibitem{chen2020generating}
Chen, Z., Song, Y., Chang, T.H., Wan, X.: Generating radiology reports via memory-driven transformer. arXiv preprint arXiv:2010.16056  (2020)

\bibitem{cornia2020meshed}
Cornia, M., Stefanini, M., Baraldi, L., Cucchiara, R.: Meshed-memory transformer for image captioning. In: Proc. IEEE Conf. Comput. Vis. Pattern Recogn. pp. 10578--10587 (2020)

\bibitem{demner2016preparing}
Demner-Fushman, D., Kohli, M.D., Rosenman, M.B., Shooshan, S.E., Rodriguez, L., Antani, S., Thoma, G.R., McDonald, C.J.: Preparing a collection of radiology examinations for distribution and retrieval. J. Am. Med. Inform. Assoc.  \textbf{23}(2),  304--310 (2016)

\bibitem{denkowski2011meteor}
Denkowski, M., Lavie, A.: Meteor 1.3: Automatic metric for reliable optimization and evaluation of machine translation systems. In: Proceedings of the sixth workshop on statistical machine translation. pp. 85--91 (2011)

\bibitem{feng2017deep}
Feng, J., Zhou, Z.H.: Deep miml network. In: AAAI Conf. Artif. Intell. vol.~31 (2017)

\bibitem{gamper2021multiple}
Gamper, J., Rajpoot, N.: Multiple instance captioning: Learning representations from histopathology textbooks and articles. In: Proc. IEEE Conf. Comput. Vis. Pattern Recogn. pp. 16549--16559 (2021)

\bibitem{guevara2023caption}
Guevara, B.C., Marini, N., Marchesin, S., Aswolinskiy, W., Schlimbach, R.J., Podareanu, D., Ciompi, F.: Caption generation from histopathology whole-slide images using pre-trained transformers. In: Medical Imaging with Deep Learning, short paper track (2023)

\bibitem{huang2023visual}
Huang, Z., Bianchi, F., Yuksekgonul, M., Montine, T.J., Zou, J.: A visual--language foundation model for pathology image analysis using medical twitter. Nat. Med.  \textbf{29}(9),  2307--2316 (2023)

\bibitem{ilse2018attention}
Ilse, M., Tomczak, J., Welling, M.: Attention-based deep multiple instance learning. In: Proc. Int. Conf. Mach. Learn. pp. 2127--2136. PMLR (2018)

\bibitem{johnson2019mimic}
Johnson, A.E., Pollard, T.J., Berkowitz, S.J., Greenbaum, N.R., Lungren, M.P., Deng, C.y., Mark, R.G., Horng, S.: Mimic-cxr, a de-identified publicly available database of chest radiographs with free-text reports. Sci. Data  \textbf{6}(1), ~1--8 (2019)

\bibitem{khened2021generalized}
Khened, M., Kori, A., Rajkumar, H., Krishnamurthi, G., Srinivasan, B.: A generalized deep learning framework for whole-slide image segmentation and analysis. Scientific reports  \textbf{11}(1),  11579 (2021)

\bibitem{li2021dual}
Li, B., Li, Y., Eliceiri, K.W.: Dual-stream multiple instance learning network for whole slide image classification with self-supervised contrastive learning. In: Proc. IEEE Conf. Comput. Vis. Pattern Recogn. pp. 14318--14328 (2021)

\bibitem{lin2004rouge}
Lin, C.Y.: Rouge: A package for automatic evaluation of summaries. In: Text summarization branches out. pp. 74--81 (2004)

\bibitem{lu2023visual}
Lu, M.Y., Chen, B., Zhang, A., Williamson, D.F., Chen, R.J., Ding, T., Le, L.P., Chuang, Y.S., Mahmood, F.: Visual language pretrained multiple instance zero-shot transfer for histopathology images. In: Proc. IEEE Conf. Comput. Vis. Pattern Recogn. pp. 19764--19775 (2023)

\bibitem{nicolson2023improving}
Nicolson, A., Dowling, J., Koopman, B.: Improving chest x-ray report generation by leveraging warm starting. Artificial intelligence in medicine  \textbf{144},  102633 (2023)

\bibitem{oquab2023dinov2}
Oquab, M., Darcet, T., Moutakanni, T., Vo, H., Szafraniec, M., Khalidov, V., Fernandez, P., Haziza, D., Massa, F., El-Nouby, A., et~al.: Dinov2: Learning robust visual features without supervision. arXiv preprint arXiv:2304.07193  (2023)

\bibitem{papineni2002bleu}
Papineni, K., Roukos, S., Ward, T., Zhu, W.J.: Bleu: a method for automatic evaluation of machine translation. In: Proceedings of the 40th annual meeting of the Association for Computational Linguistics. pp. 311--318 (2002)

\bibitem{sengupta2023automatic}
Sengupta, S., Brown, D.E.: Automatic report generation for histopathology images using pre-trained vision transformers. arXiv preprint arXiv:2311.06176  (2023)

\bibitem{shao2021transmil}
Shao, Z., Bian, H., Chen, Y., Wang, Y., Zhang, J., Ji, X., et~al.: Transmil: Transformer based correlated multiple instance learning for whole slide image classification. Proc. Adv. Neural Inf. Process. Syst.  \textbf{34},  2136--2147 (2021)

\bibitem{tomczak2015review}
Tomczak, K., Czerwi{\'n}ska, P., Wiznerowicz, M.: Review the cancer genome atlas (tcga): an immeasurable source of knowledge. Contemporary Oncology/Wsp{\'o}{\l}czesna Onkologia  \textbf{2015}(1),  68--77 (2015)

\bibitem{vaswani2017attention}
Vaswani, A., Shazeer, N., Parmar, N., Uszkoreit, J., Jones, L., Gomez, A.N., Kaiser, {\L}., Polosukhin, I.: Attention is all you need. Adv Neural Inf Process Syst  \textbf{30} (2017)

\bibitem{veta2019predicting}
Veta, M., Heng, Y.J., Stathonikos, N., Bejnordi, B.E., Beca, F., Wollmann, T., Rohr, K., Shah, M.A., Wang, D., Rousson, M., et~al.: Predicting breast tumor proliferation from whole-slide images: the tupac16 challenge. Med. Image Anal.  \textbf{54},  111--121 (2019)

\bibitem{vinyals2015show}
Vinyals, O., Toshev, A., Bengio, S., Erhan, D.: Show and tell: A neural image caption generator. In: Proc. IEEE Conf. Comput. Vis. Pattern Recogn. pp. 3156--3164 (2015)

\bibitem{wang2023controllable}
Wang, N., Xie, J., Wu, J., Jia, M., Li, L.: Controllable image captioning via prompting. In: AAAI Conf. Artif. Intell. vol.~37, pp. 2617--2625 (2023)

\bibitem{wang2022transformer}
Wang, X., Yang, S., Zhang, J., Wang, M., Zhang, J., Yang, W., Huang, J., Han, X.: Transformer-based unsupervised contrastive learning for histopathological image classification. Med. Image Anal.  \textbf{81},  102559 (2022)

\bibitem{zhang2022dtfd}
Zhang, H., Meng, Y., Zhao, Y., Qiao, Y., Yang, X., Coupland, S.E., Zheng, Y.: Dtfd-mil: Double-tier feature distillation multiple instance learning for histopathology whole slide image classification. In: Proc. IEEE Conf. Comput. Vis. Pattern Recogn. pp. 18802--18812 (2022)

\end{thebibliography}

\section*{Supplementary Materials}

\begin{figure}[htbp]
  \centering
  \includegraphics[width=\textwidth]{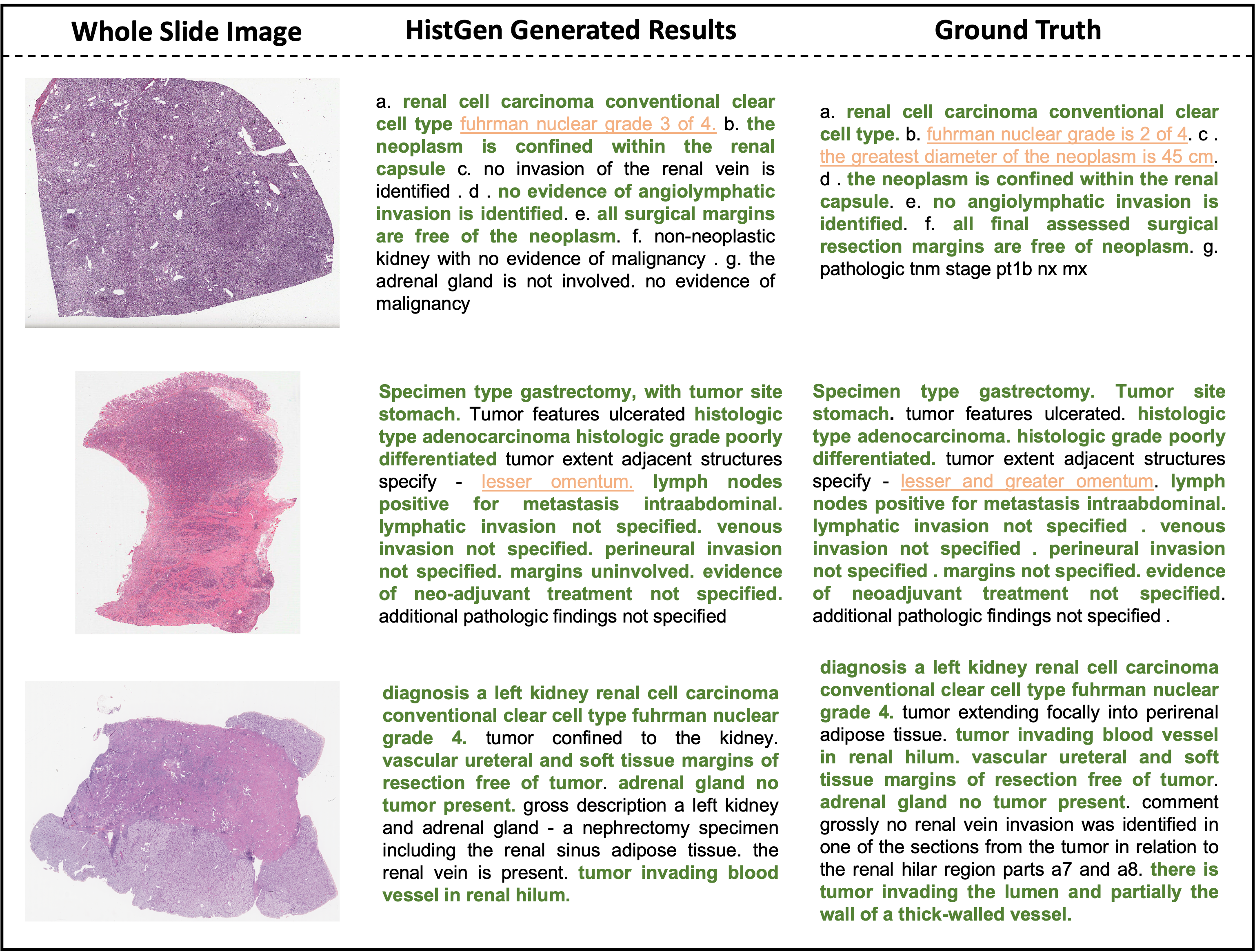}
  \caption{Qualitative analysis of the proposed HistGen model. Words highlighted in bold green indicate alignment between our model’s generated results and the ground truth. Conversely, words underlined in orange represent diagnostic details that our model fails to capture. The first two examples highlight the superior captioning capability of our model, with it accurately diagnosing provided WSIs. The diagnoses closely align with the ground truths, differing only in minor, non-critical aspects. In the third example, our model successfully makes the correct prediction, despite the absence of the detailed context present in the ground truth.}
  \label{fig:qualitative}
\end{figure}

\begingroup
\setlength{\tabcolsep}{4pt} 
\renewcommand{\arraystretch}{1.5} 
\begin{table}[h]
\centering
\caption{Ablation study on region size of local-global hierarchical encoder. For WSI report generation, we adjust the region size within the local-global hierarchical encoder, ranging from 64 to 512, to identify the optimal setting for this task. The results reveal that the impact of the region size choice is relatively minor. This phenomenon could be attributed to our curated dataset's insensitivity to this particular hyperparameter. Instead, it appears that the effectiveness of the task hinges more on our model's region-to-slide learning mechanism.}
\scalebox{0.8}{\begin{tabular}{lccccccc}
\toprule
\multicolumn{1}{c}{\multirow{2}{*}{\textbf{Methods}}}  & \multicolumn{6}{c}{\textbf{Metric}} \\ 
\cline{2-7}
& BLEU-1 & BLEU-2 & BLEU-3 & BLEU-4 & METEOR & ROUGE-L \\ 
\midrule
\textbf{Ours w/ Region Size 64} &0.411 &0.295 & 0.228& 0.184&0.181 &0.343 \\
\hline
\rowcolor{red!10!white}
\textbf{Ours w/ Region Size 96} & \textbf{0.413}& \textbf{0.297}& \textbf{0.229}& \textbf{0.184}& \textbf{0.182 }&\textbf{0.344} \\
\hline
\textbf{Ours w/ Region Size 128} & 0.405& 0.290& 0.224&0.180 &0.178  &0.338 \\
\hline
\textbf{Ours w/ Region Size 256} & 0.408 & 0.291 & 0.224 & 0.179 & 0.179 & 0.338 \\
\hline
\textbf{Ours w/ Region Size 384} & 0.411 & 0.294 & 0.227 & 0.182 & 0.181 & 0.341 \\
\hline
\textbf{Ours w/ Region Size 512} & 0.406  & 0.291 & 0.226 & 0.182 &0.179 & 0.341 \\
\bottomrule
\end{tabular}}
\label{tab:region}
\end{table}
\endgroup

\begingroup
\setlength{\tabcolsep}{4pt} 
\renewcommand{\arraystretch}{1.4} 
\begin{table}[htbp]
\centering
\caption{Composition of the curated WSI-Report dataset. The dataset structure is detailed as follows, with each subset corresponding to a distinct disease type. Raw reports and more information on different subsets could be sourced from \href{https://portal.gdc.cancer.gov}{the TCGA platform}. Notably, to guarantee each WSI corresponds to the report description, cases where one report matches multiple WSIs are excluded.}
\begin{adjustbox}{width=\textwidth}
\scalebox{0.8}{
\begin{tabular}{c|cccccccc}
\toprule
\toprule
Cancer Type & BRCA & UCEC & KIRC & THCA & LGG  & LUAD & HNSC & LUSC \\ \hline
Number      & 999  & 503  & 501  & 486  & 444  & 444  & 442  & 442  \\ \midrule\midrule
Cancer Type & COAD & PRAD & BLCA & STAD & LIHC & KIRP & CESC & GBM  \\ \hline
Number      & 412  & 368  & 353  & 337  & 329  & 265  & 251  & 245  \\ \midrule\midrule
Cancer Type & SARC  & PAAD  & PCPG  & READ  & ESCA & THYM  & KICH & SKCM \\ \hline
Number      & 243 & 174 & 171 & 156 & 117 & 113  & 108  &  97 \\ \midrule\midrule
Cancer Type & TGCT & MESO & UVM & OV & UCS & ACC & DLBC & CHOL \\ \hline
Number      & 87  & 67 & 65 & 62 & 56 & 54 & 43 & 37 \\ \bottomrule\bottomrule
\end{tabular}}
\end{adjustbox}
\label{tab:dataset}
\end{table}
\endgroup


\begin{figure}[htbp]
  \centering
  \includegraphics[width=11cm]{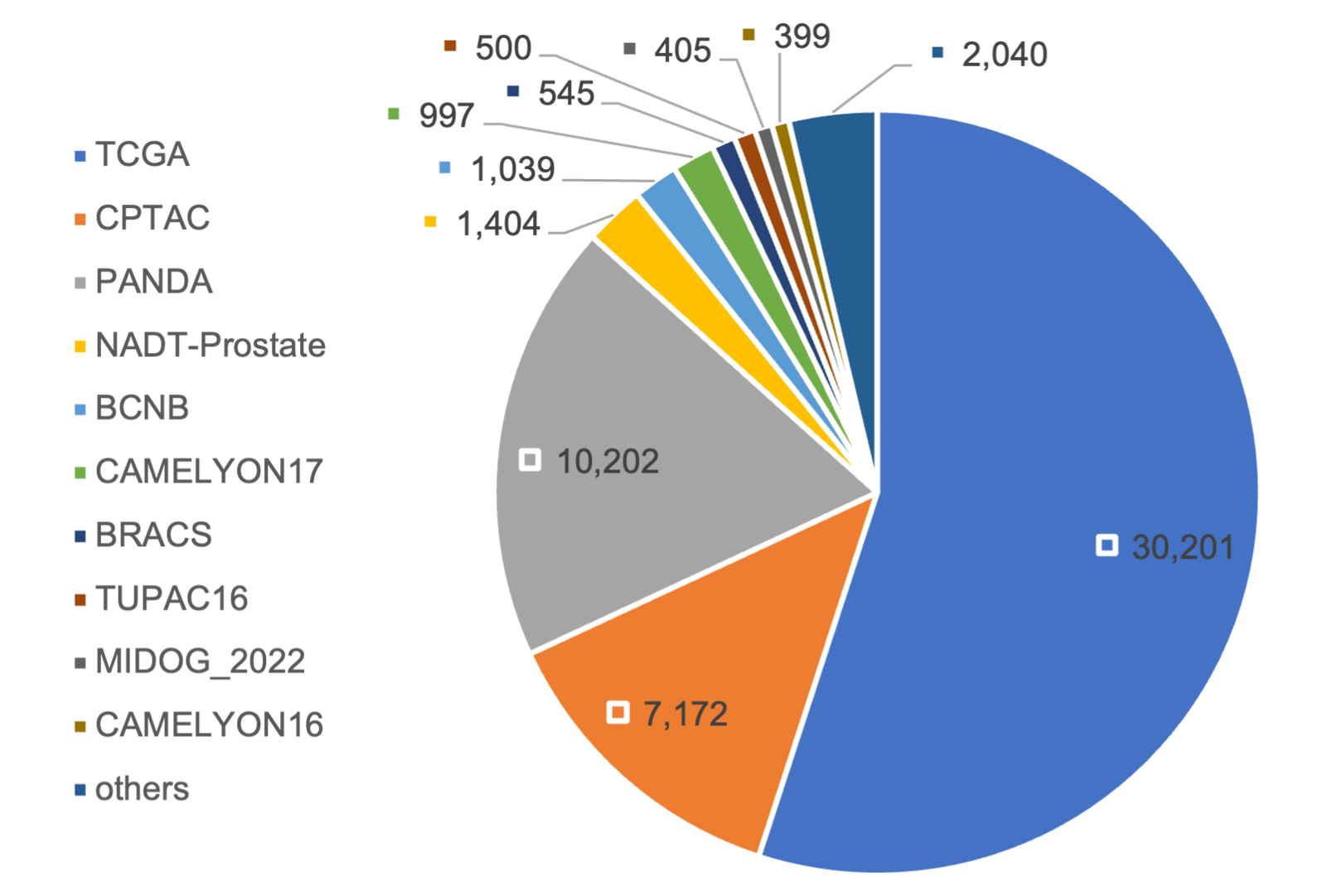}
  \caption{WSI distribution for DINOv2 ViT-L feature extractor pre-training. We have collected over 30 different pathology datasets containing over 60 primary sites. Patches are extracted from whole slide images at level $0$, with dimensions of $512\times 512$. These patches are subsequently resized to $224\times 224$ for pre-training the feature extractor. This figure shows the details of our collected WSIs.}
  \label{fig:WSI_distribution}
\end{figure}

\begin{figure}[htbp]
  \centering
  \includegraphics[width=\textwidth]{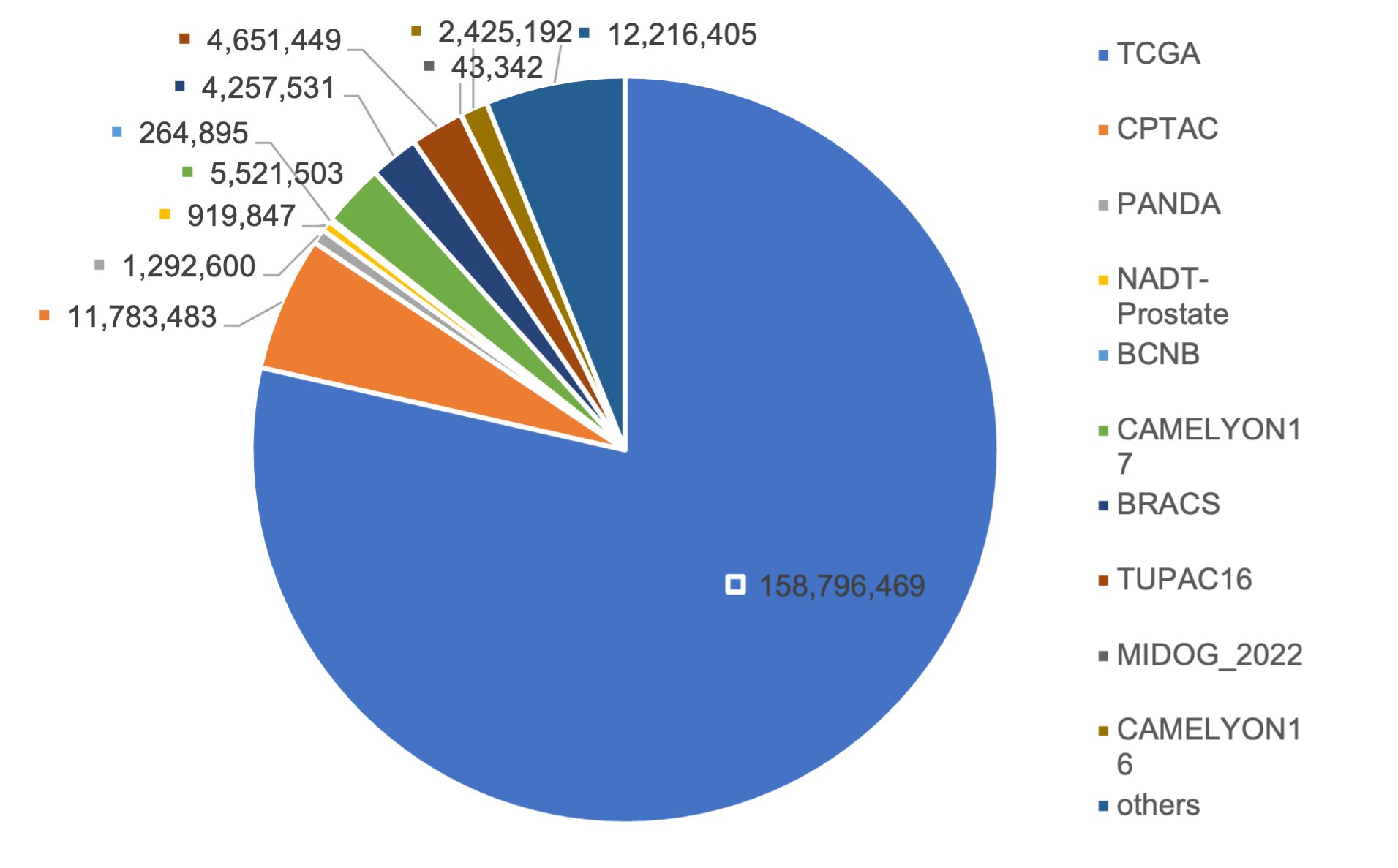}
  \caption{Patch distribution for DINOv2 ViT-L feature extractor pre-training.}
  \label{fig:patch_distribution}
\end{figure}

\end{document}